\documentclass[runningheads]{llncs}
\usepackage{graphicx}

\usepackage{tikz}
\usepackage{comment}
\usepackage{amsmath,amssymb} 
\usepackage{color}
\usepackage{bm} 
\usepackage[ruled]{algorithm2e} 
\usepackage{booktabs} 
\usepackage{multirow}
\usepackage{marvosym}
\usepackage{subfig,graphicx}
\usepackage{hyperref}
\hypersetup{
    colorlinks=true,
    linkcolor=red,
    filecolor=magenta,      
    urlcolor=magenta
    }

\usepackage[accsupp]{axessibility}  

\begin{document}
\pagestyle{headings}
\mainmatter
\def\ECCVSubNumber{1179}  

\title{Anti-Retroactive Interference for \protect\\ Lifelong Learning} 


\titlerunning{Anti-Retroactive Interference for Lifelong Learning}
%
\author{Runqi Wang \inst{1} \and
Yuxiang Bao\inst{1} \and
Baochang Zhang\inst{1}\thanks{ Corresponding author.}  \and Jianzhuang Liu\inst{2} \and Wentao Zhu\inst{3} \and Guodong Guo\inst{4}}
\authorrunning{R. Wang et al.}
%
\institute{Beihang University \and
Huawei Noah's Ark Lab \and Kuaishou Technology  \and Institute of Deep Learning, Baidu Research\\
\email{\{runqiwang,bczhang\}@buaa.edu.cn}}
\maketitle

\begin{abstract}
Humans can continuously learn new knowledge. However, machine learning models suffer from drastic dropping in performance on previous tasks after learning new tasks. Cognitive science points out that the competition of similar knowledge is an important cause of forgetting. In this paper, we design a paradigm for lifelong learning based on meta-learning and associative mechanism of the brain. It tackles the problem from two aspects: extracting knowledge and memorizing knowledge. First, we disrupt the sample's background distribution through a background attack, which strengthens the model to extract the key features of each task. Second, according to the similarity between incremental knowledge and base knowledge, we design an adaptive fusion of incremental knowledge, which helps the model allocate capacity to the knowledge of different difficulties. It is theoretically analyzed that the proposed learning paradigm can make the models of different tasks converge to the same optimum. The proposed method is validated on the MNIST, CIFAR100, CUB200 and ImageNet100 datasets. The code is available at \textit{\url{https://github.com/bhrqw/ARI}}.

\end{abstract}

\section{Introduction}
\label{sec:intro}

A standard benchmark for success in artificial intelligence is the ability to emulate human learning. However, at the current stage, the machine does not really understand what it has learned. It may just do rote memorization, which overlooks a critical characteristic of human learning: being robust to changing tasks and sequential experience. Future learning machines should be able to adapt to the ever-changing world. They should continuously learn new tasks without forgetting previously learned ones. Although many learning paradigms have been proposed, such as lifelong learning (LLL)~\cite{aljundi2018memory,rajasegaran2020itaml}, these problems have not been addressed well. Many researchers are brute-force and idealized in the construction of model training. 
In pedagogy and psychology, human learning and the cognitive process have been widely discussed, among which there are many theories worthy of reference. The learning process of new tasks results in catastrophic forgetting~\cite{hadsell2020embracing} of previous knowledge due to \textbf{retroactive interference}~\cite{sternberg2012cognitive}, which means that the content of later learning competes with the previous memory and interferes with the previous learning. This kind of competition causes confusion and forgetting of knowledge. This problem can be solved by capturing critical points of knowledge and removing redundant content to avoid the competition of knowledge, which is termed filter efficiency~\cite{vogel2005neural} in pedagogy. In the computer vision task of classification, different categories of images might have the same or similar backgrounds, such as a bicycle and a dog on a lawn. Machine learning models may mistake lawn features for bicycle features or dog features, which creates unnecessary memory competition in learning new knowledge.

\begin{figure}[t]
    \centering
    \includegraphics[width=0.6\linewidth]{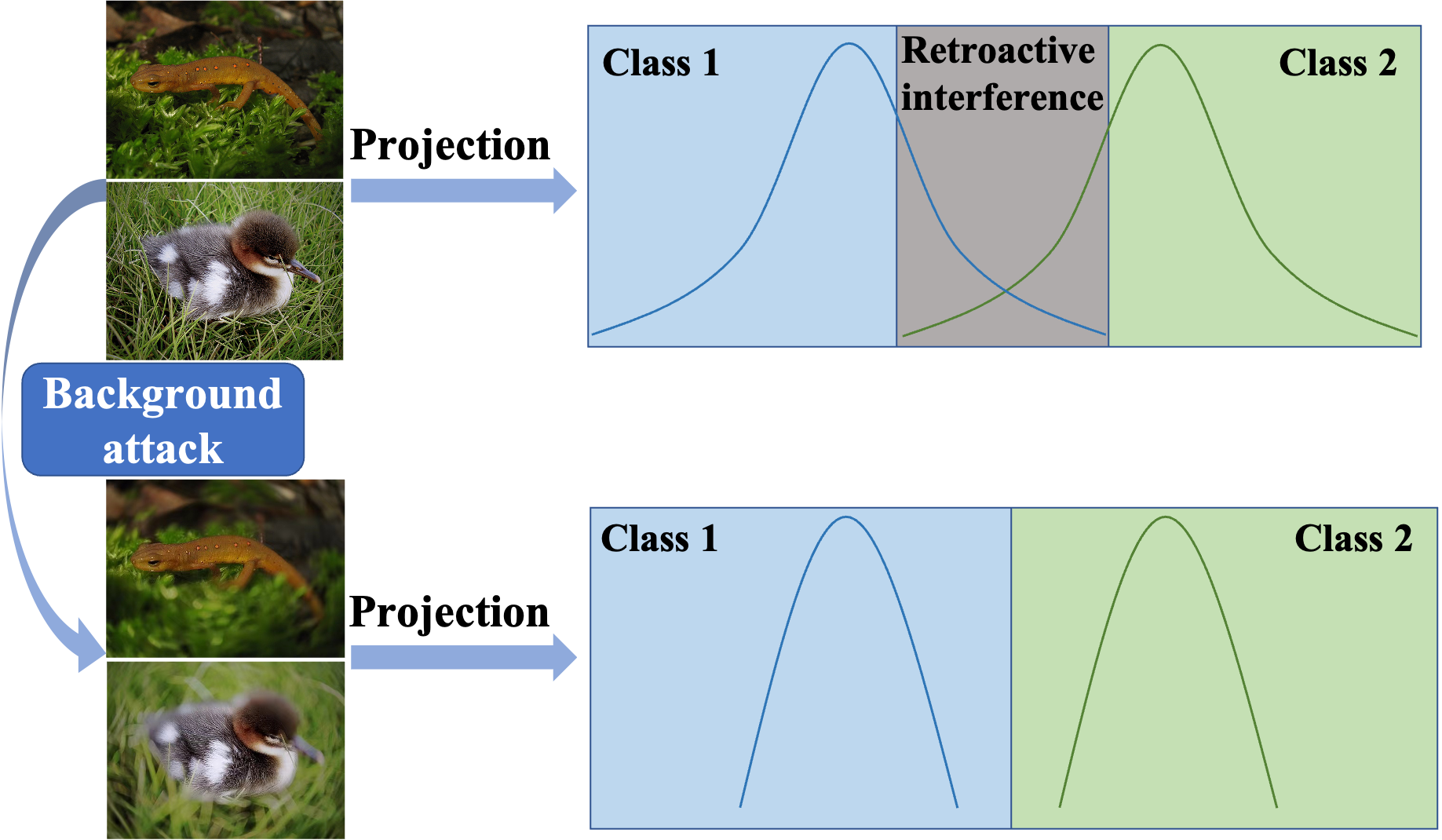}
    \caption{In lifelong learning, the knowledge in the present stage competes with the previous memory and interferes with the previous learning, especially when the knowledge is similar. Therefore, we attack similar contents in both tasks to change the data distribution and avoid retroactive interference.}
    \label{fig:feature}
\end{figure}

In order to solve the problem of forgetting in the process of learning new tasks in a deep learning model, we 
propose a lifelong learning paradigm based on meta-learning and associative learning. We divide the model into two stages: extracting intra-class features and fusing inter-class features. In the first stage, we hope to avoid retroactive interference by reducing the competition between old and new knowledge. We need to accurately capture the critical knowledge of new tasks and focus on learning it, which can effectively avoid confusion of knowledge. 
In this way, incremental knowledge can complement rather than compete with existing knowledge. It is an anthropomorphic process that associates the original images with the foreground of the images, i.e., only learning the critical knowledge of the new task as a complement to the knowledge base. In this way, model information redundancy can be avoided, which is consistent with the machine learning theory in~\cite{mrmr}. In order to realize this idea, we present a background attack method to attack the samples adversarially. Through the spatial attention mechanism, the importance map of the image can be obtained. We believe that areas of low importance level in an image do not belong to the necessary information of its class, which may cause information redundancy and competition between classes as shown in Fig. \ref{fig:feature}. Therefore, we carry out an adversarial attack on non-critical areas, ($i.e.,$ the background) and blur the data distribution in these areas, thus weakening the model's learning of unimportant information. 

In the second stage, we combine the existing model with the model just learned. It is different from conventional incremental learning that updates the pre-trained model directly, which is easy to cause catastrophic damage to the model's weight distribution. We organize the knowledge to learn into different tasks, just like the chapters of a textbook. Each task is learned separately, and an independent model is outputted. Specifically, when learning a new task, a small number of samples are extracted from previous tasks for review, and then the models corresponding to these tasks are fused, which is consistent with Ausubel's theory~\cite{ausubel1961role} that points out that the most important thing in learning is whether the knowledge learned can form a system, $i.e.$, to complete the deduction of knowledge from the individual to the whole. Following this process, we chose a meta-training based method to generate models, which will be described in Sec.~\ref{sec:method}. To this end, we propose a novel task-specific fusion method, and show that our training process can ensure that these different models are converged to a common optimal one to reduce the information loss. Our contributions are summarized as follows:

\begin{itemize}
\item 
We combine the adversarial attack with meta-learning to extract features. The adversarial attack is performed on the image background to reinforce the model's attention to critical features.
\item
Based on human cognition, a new lifelong learning paradigm, \textit{Anti-Retroactive Interference for lifelong learning} (ARI), is established to ensure that the machine learning model can integrate incremental knowledge more effectively. It is analyzed that the fusion method in ARI can ensure that the task-specific models are converged to the same optimal model to reduce the information loss caused by fusion.

\item
The proposed method is validated on the MNIST, CIFAR100, CUB200 and ImageNet100 datasets, and state-of-the-art results are obtained on all the benchmarks. 
\end{itemize}

\section{Related Work}
\label{sec:related work}

\subsection{Lifelong Learning }

So far, lifelong learning methods can be divided into three groups. The first one is based on regularization. LwF~\cite{li2017learning} preserves the ancient knowledge by adding a distillation loss. In addition, the distillation loss is implemented by~\cite{rebuffi2017icarl,castro2018end,zhao2020maintaining} to reduce forgetting. \cite{wu2019large,zhao2020maintaining} propose bias correction strategies whereby the model can perform equally well on current and older classes by re-balancing the ﬁnal fully-connected layer. 
EWC~\cite{kirkpatrick2017overcoming} computes synaptic importance offline by calculating a Fisher information matrix. E-MAS-SDC proposed by \cite{yu2020semantic} estimates the drift of previous tasks during the training of new tasks to make semantic drift compensation. RRR in \cite{Ebrahimi2021RememberingFT} tries to save the correct attentions of previous images to avoid the attentions being affected by other tasks.
The second group is about expanding the model with progressive learning  and designing binary masks that directly map each task to the corresponding model architecture. MARK~\cite{hurtado2021optimizing} keeps a set of shared weights among tasks. These shared weights are envisioned as a common knowledge base used to learn new tasks and enriched with new knowledge as the model learns new tasks. In~\cite{abati2020conditional}, each convolutional layer is equipped with task-specific gating modules, selecting specific ﬁlters for a given task. The shortcomings of these methods are the extra model complexity and the need for a practical scheme to calculate the mask precisely. 
The third group is replay based and it gets popular recently. Replay based approaches are ideally suitable for lifelong learning in which tasks are added in turn. iTAML~\cite{rajasegaran2020itaml} introduces a meta-learning approach that seeks to maintain an equilibrium between all the encountered tasks, in the sense that it is unbiased towards class samples of majority and simultaneously minimizes forgetting.

\subsection{Adversarial Training}
Though the success of deep learning models has been demonstrated on various computer vision tasks, they are sensitive to adversarial attacks~\cite{goodfellow2016deep}. An imperceptible perturbation added to inputs may cause undesirable outputs. The Fast Gradient Sign Method (FGSM) is proposed in \cite{goodfellow2014explaining} to generate adversarial examples with a single gradient step. 
To defend the attacks, many methods have been proposed to defend against them. The most common method is adversarial training \cite{na2017cascade,kurakin2016adversarial,tramer2018ensemble} with adversarial examples added to the training data. In this paper, we introduce adversarial training to the meta-learning process to obtain a robust model that can extract good features from very few available samples.

\section{Proposed Method}
\label{sec:method}

We adopt a task-incremental learning setup where the model continuously learns new tasks, each containing a fixded number of novel classes. During the training process of task $n$, we have access to $\mathbb{M}_{n-1}$ and $\mathbb{D}_n$ where $\mathbb{M}_{n-1}$ is an exampler memory containing a small number of samples for old tasks, and $\mathbb{D}_n$ is the training data for task $n$, which contains pairs $(\mathbf{x}_i, y_i)$, with $\mathbf{x}_i$ being an image of class $y_i \in R_n$. Using $\mathbb{M}_{n-1}$ to train task $n$ is a form of meta-learning. We define the set of classes on task $n$ as $R_n = \{ r_{n,1}, r_{n,2}, ..., r_{n,m} \}$, where $r_{n,1}$ is the first class in task $n$, and $m$ is the number of classes in task $n$. Different tasks do not contain the same class: $R_t \cap R_s = \varnothing, t \neq s$. After learning all the tasks, we evaluate the learned model on all tasks $R = \cup_i R_i$.

\subsection{Extracting Intra-Class Features}

\begin{algorithm}
	\caption{Associative learning with background attack}
	\label{alg:ba}
	\textbf{Input:} Training data $\mathbf{x}$;\\ \textbf{Hyper-parameters:} Epoch number $S$, $\varepsilon =  \frac{8}{255}$; \\
	Initialize model parameters $\bm{\theta}$ ; \\
	\textbf{Output:} The network model; \\
	$\#$ Train an architecture for $S$ epochs: \\
	 $t=0$ ; \\
	\While{$(t\leq S)$}
        {$\#$ First inference:\\
        Input $\mathbf{x}$;\\
        According to~\cite{woo2018cbam}, calculate the spatial attention $\mathbf{A}$; \\
        Return $\mathbf{A}$;\\
        $\#$Back propagation:\\
        According to Eq.~\ref{eq:attacks}, calculate $\mathbf{x'}$;\\
        $\#$ Second inference:\\
        Input $\mathbf{x'}$;\\
        $\#$Back propagation:\\
        Update parameters $\bm{\theta}$;\\
        $t \leftarrow t + 1$.}
\end{algorithm}

Lifelong learning requires the model to retain previous knowledge and learn new knowledge. However, if the previous and new knowledge have similar characteristics, it is easy to cause forgetting. Data are labeled for different classes according to different object features, but the background information is ignored, which may mislead the model's incremental learning. In order to eliminate similar characteristics between different classes and prevent retroactive interference, we design associative learning with background attack. This approach involves two processes. In the first process, the model learns from the original image to obtain the background region and conduct adversarial attack on it. This attack can disturb the distribution of the background and strengthen the feature extraction on the critical region of the image. In the second process, the model is trained with the attacked images. This approach associates the objects with different backgrounds, which avoids the negative effect of background on few-shot learning. Therefore, the model can effectively avoid  over-fitting by associative learning.

In adversarial training, we need to add perturbation to the images, which can increase the robustness of the model. However, now we use a background mask $\mathbf{B}$ to guide the model to attack the background regions of the images. The mask $\mathbf{B}$ has three forms:
\begin{equation}
\begin{aligned}
    \mathbf{B} = 1-\mathbf{A},~~ \mathbf{B} = 1-\mathbf{A}\circ\mathbf{A},~~\mathbf{B} = \frac{1}{\mathbf{A}},
\end{aligned}
\label{eq:B}
\end{equation}
where $\mathbf{A} \in \mathbb{R}^{s \times s}$ denotes the spatial attention obtained by~\cite{woo2018cbam}, $\circ$ denotes the Hadamard product, $\mathbf{B} \in \mathbb{R}^{s \times s}$ is the mask for focusing on the background, and $s \times s$ is the size of the image. In order to widen the distance between important and unimportant information in the attention and guide the background attack, we use the three forms of $\mathbf{B}$ in Eq.~\ref{eq:B}, making the unimportant regions (corresponding more to the background) prominent. Therefore, the attack guided by $\mathbf{B}$ tends to be more selective on the background.

We formulate the background attack model as:
\begin{equation}
    \begin{aligned}
    \mathbf{x'} &= \mathbf{x} + \mathbf{B}\circ\bm{\zeta} = \mathbf{x} + \mathbf{B}\circ (\varepsilon \mathbf{sgn}(\nabla_\mathbf{x} G(\bm{\theta}, \mathbf{x}, y))),\\
    \end{aligned}
    \label{eq:attacks}
\end{equation}
where $\mathbf{x} \in \mathbb{R}^{s \times s}$ is the clean input and $\mathbf{x'}$ is the adversarial counterpart. $\bm{\zeta}$ denotes the global perturbation of the clean input $\mathbf{x}$ which is designed based on~\cite{goodfellow2014explaining}. $y$ denotes the label of the input $\mathbf{x}$. $\varepsilon$ is the perturbation bound, $\bm{\theta}$ denotes the parameters of the deep model, and $G$ is the cross-entropy function.

The algorithm of the associative learning with adversarial background attack is listed in Algorithm~\ref{alg:ba}, which associates clean input $\mathbf{x}$ with various adversarial inputs $\mathbf{x'}$. After the adversarial training with $\mathbf{x'}$, the model learns to be robust to the distribution shift~\cite{zhang2021delving} of background and thus can focus more on the foreground (object) features, reducing forgetting as shown in Fig.~\ref{fig:feature}. Experimental verification is shown in Sec.~\ref{sec:4.4}.

\subsection{Generating and Fusing Task-Speciﬁc Models}

\begin{figure}[t]
    \centering
    \includegraphics[width=1\linewidth]{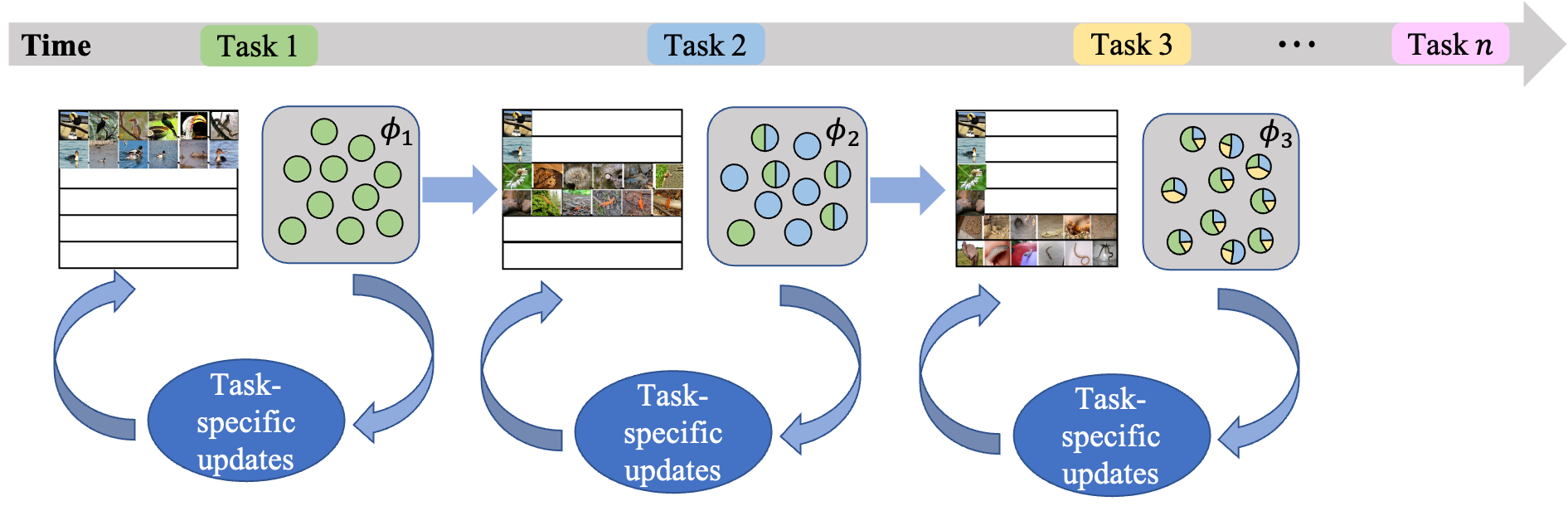}
    \caption{We design a serial learning structure for lifelong learning. A small-scale rehearsal memory of the previous tasks is also used to fine-tune the new model to adapt to the new task.}
    \label{fig:incremental}
\end{figure}

\begin{figure*}[t]
    \centering
    \includegraphics[width=1\linewidth]{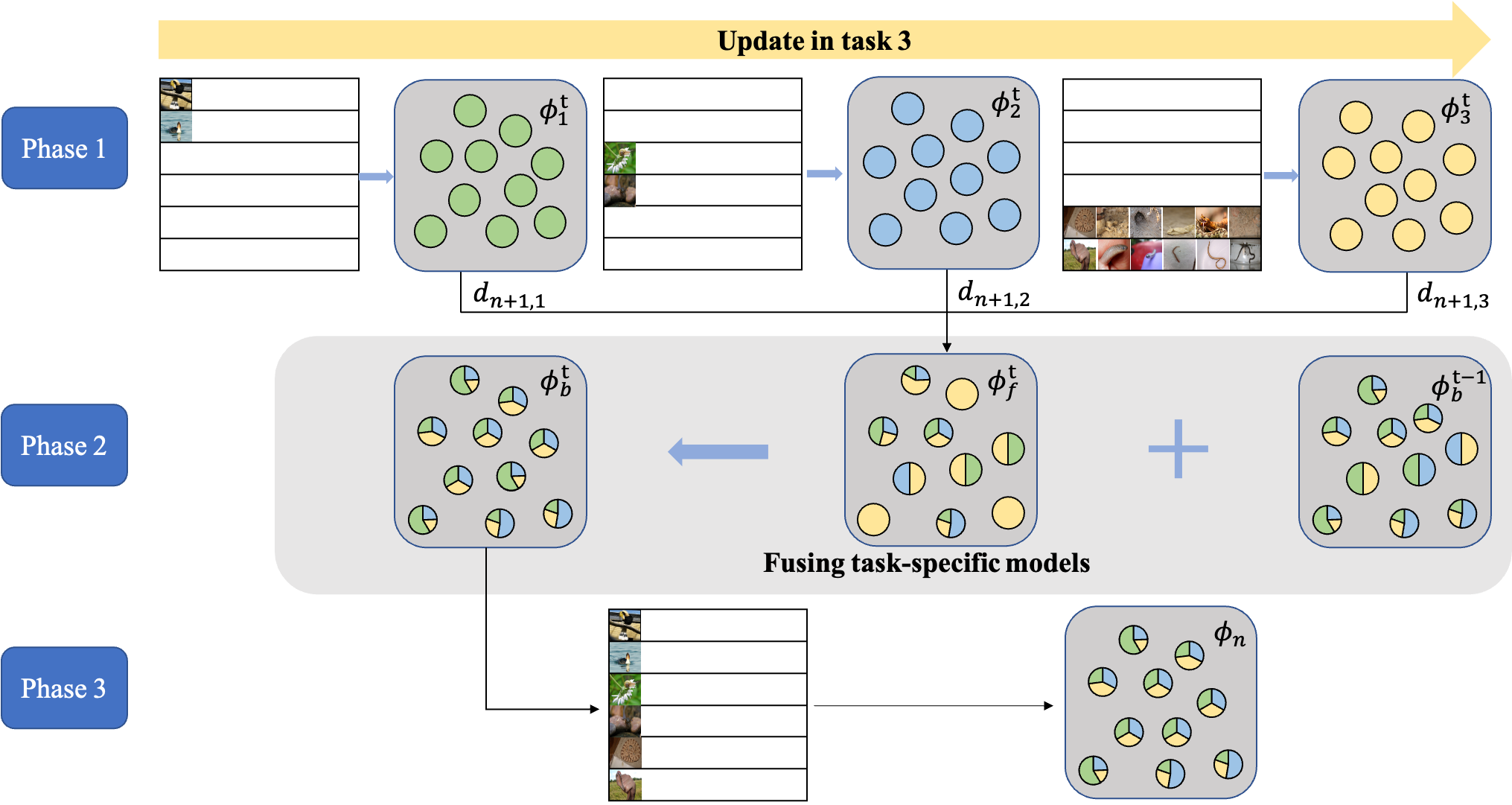}
    \caption{Taking task 3 for example, the task-specific updating is divided into three phases. In phase 1, task-specific model training is carried out. It is noteworthy that only a small amount of previous task samples are used in the current training. In phase 2, task-specific models are fused. In phase 3, meta-training is performed on the fused model to obtain the incremental base model of task 3.}
    \label{fig:process}
\end{figure*}

The adversarial images after the background attack are used as input to participate in training. Lifelong learning is a scenario in which tasks are entered serially. The base model should contain information about all learned tasks after learning a new task, as shown in Fig.~\ref{fig:incremental}, in which $\bm{\phi}_n$ denotes the base model after learning task $n$. The process of learning a new task is embedded in the task-specific updating. When updating in a new task, our meta-learning approach involves three phases: (1) generating task-specific models for all the seen tasks, (2) fusing the task-specific models into the base model, and (3) meta-training the base model, as shown in Fig.~\ref{fig:process}.

\begin{algorithm}
	\caption{Training in task $n$}
	\label{alg:tst}
	\textbf{Input:} $\mathbb{D}_n$, $\mathbb{M}_{n-1}$, $\phi_{n-1}$; \\ \textbf{Hyper-parameters:} task number $n$, epoch number $S$, image number $J_i$ of task $i$, $i \in [1,n]$;\\
	\textbf{Output:} The base model $\phi_{n}$; \\
	$\#$ Train an architecture for $S$ epochs; \\
    $\phi_b^0= \phi_{n-1}$, $t=1$;\\
	\While{$(t\leq S)$}
        {
        \For{$i = 1\ \mathbf{to}\ n$}
        {$\left\{\hat{y}_j^i\right\}_{j=1}^{J_i} \leftarrow \phi_{b}^{t-1}\left(\left\{\mathbf{x}_j^i\right\}_{j=1}^{J_i}\right)$;\\
        $loss \leftarrow \mathbf{Eq.}~\ref{eq:loss}$;
        }
        $\phi_i^t \leftarrow \mathbf{Optimizer}(\phi_{b}^{t-1}, loss)$; \\
        $\phi_{f}^t \leftarrow \mathbf{Fusion} [\phi_1^t, ... , \phi_n^t, \phi_b^{t-1}]$;\\
        $\phi_b^t \leftarrow \gamma\phi_{f}^t + (1-\gamma)\phi_b^{t-1} $;\\
        $t \leftarrow t + 1$;
        }
        $\phi_{n} \leftarrow$ \textbf{Meta train} $(\phi_b^S)$.
\end{algorithm}

\textbf{Generating task-specific models.}
We randomly sample a mini-batch $\mathbb{B}_n = \{(\mathbf{x}_k, y_k)\}_{k=1}^{K_n}$ from the current task $n$ training data $\mathbb{D}_n$ and the memory bank $\mathbb{M}_{n-1}$, which contains a few samples for old tasks. $\mathbf{x}_k$ and $y_k$ are the training images and their labels, respectively, and $K_n$ is the image number of the batch. Therefore, the mini-batch of data for task-specific updates, as shown in Fig.~\ref{fig:process}, is represented as: 
\begin{equation}
   \mathbb{B}_n \thicksim  \mathbb{D}_n \cup \mathbb{M}_{n-1}.
\end{equation}

We sample the training data according to the tasks to construct $\mathbb{B}^i_\mu = \{(\mathbf{x}_j^i, y_j^i)\}_{j=1}^{J_i}$ for training the task-specific models $\phi_i$, $i\in [1, n]$, where $J_i$ is the image number of task $i$. The loss function in the task-specific updating is the binary cross-entropy loss with a regularizer from $\mathbf{dif}$, which is defined next in Eq.~\ref{eq:cor}. The binary cross-entropy is:
\begin{equation}
\begin{aligned}
    L(\phi_i(\{\mathbf{x}_j^i\}), \{y_j^i\})=&-\frac{1}{J_i} \sum_{j=1}^{J_i}( y_j^i \cdot \log \left(\phi_i \left(\mathbf{x}^i_j \right)\right) +\left(1-y_j^i\right) \cdot \log (1-\log (\phi_i (\mathbf{x}_j^i)))).
\end{aligned}
\label{eq:BCEloss}
\end{equation}
This helps to obtain task-specific models $\phi_i$
, thus providing a better estimate for gradient updates in the current task-specific training (described next) to obtain a base model. The training process of the specific tasks, $i.e.$, phase 1 in Fig.~\ref{fig:process}, is shown in the for-loop of Algorithm~\ref{alg:tst}, which generates $n$ independent models. In Algorithm~\ref{alg:tst}, the Optimizer denotes some optimizer such as SGD. The function Fusion is described next (Eq.~\ref{eq:fuse}). $\phi_i^t$ is the task-specific model $i$ at epoch $t$, and $\phi_b^t$ is the base model at epoch $t$. All these models $\phi_1, ..., \phi_n, \phi_b$ have the same structure.

\textbf{Fusing task-specific models.} We combine the task-specific models $\phi_i^t$ generated during phase 1 to the base model $\phi_n$ in phase 2 of Fig.~\ref{fig:process}. We denote the set of the models at epoch $t$ as: 
\begin{equation}
    \Phi^t = \{\phi_1^t, ..., \phi_n^t, \phi_b^{t-1}\}.
\end{equation}

Due to the task-specific models being generated by different tasks, there may be large differences between their parameter values, which causes information loss in model fusion. We adopt a new strategy to use the Manhattan distance between a task-specific model and the base model as the fusion weight. When the gap between the two models is larger, the fusion weight is larger. The weight coefficients are calculated as follows. First, we define
\begin{equation}
\mathbf{dif}=\left[\begin{array}{cccc}
0 & d_{1,2} & \cdots & d_{1, n+1} \\
d_{2,1} & 0 & & d_{2, n+1} \\
\vdots & & \ddots & \vdots \\
d_{n+1, 1} & d_{n+1,2} & \cdots & 0
\end{array}\right],
\label{eq:cor}
\end{equation}
where $d_{1,2}$ denotes the Manhattan distance between $\phi_1^t$ and $\phi_2^t$, and $d_{1,n+1}$ denotes the Manhattan distance between $\phi_1^t$ and $\phi_b^{t-1}$. Considering the goal of model fusion is to minimize the differences among task-specific models and produce a fused model that performs well across tasks, we formulate the loss as two parts, the regularizer based on $\mathbf{dif}$ and the binary cross-entropy as shown in Eq.~\ref{eq:BCEloss}.
\begin{equation}
loss = L\left(\hat{y}_j^i, y_j^i\right) + \sum\limits_{a=1}^{i+1} \sum\limits_{b=1}^{i+1} |d_{a,b}|^2.
\label{eq:loss}
\end{equation}

To ensure that the sum of weights equals 1, each row of $\mathbf{dif}$ is transformed by the softmax function as:
\begin{equation}
\mathbf{dif^*}=\left[\begin{array}{cccc}
d_{1,1}^* & d_{1,2}^* & \cdots & d_{1, n+1}^* \\
d_{2,1}^* & d_{2,2}^* & & d_{2, n+1}^* \\
\vdots & & \ddots & \vdots \\
d_{n+1, 1}^* & d_{n+1,2}^* & \cdots & d_{n+1,n+1}^*
\end{array}\right].
\label{eq:dif}
\end{equation}
Finally, the fused model is formulated as:
\begin{equation}
  \phi_f^t = \sum_{i=1}^{n+1}(d_{n+1, i}^* \cdot \phi_i^t),
\label{eq:fuse} 
\end{equation}
where $\phi_{n+1}^t=\phi_b^{t-1}$. The reason we take the elements of the last row of $\mathbf{dif^*}$ as the weights is that the base model could adopt the knowledge from the task-specific models as much as possible. Thus more weights should be given to the task-specific model with a larger difference from the base model.

The fusion model is combined with $\phi_b^{t-1}$ to form a new base model $\phi_b^t$:
\begin{equation}
    \phi_b^t = \gamma\phi_{f}^t + (1-\gamma)\phi_b^{t-1},
\label{eq:phi_b}
\end{equation}
where $\gamma$ is a hyper-parameter that controls the speed of learning new information, $i.e.$, for higher $\gamma$ the model prefers to learn new information and forget the old, and with smaller $\gamma$ it learns little new knowledge.

Due to the regularization from $\mathbf{dif}$, after a sufficient number of iterations, in the sense that $t$ is large enough, the differences among the task-specific and base models $\{\phi_1^t, ..., \phi_n^t, \phi_b^{t-1}\}$ is decreasing gradually and all the models tend to have the same weights. In the supplementary material, we provide evidence to analyze that all the models converge to the same optimal weights. Moreover, an experiment is conducted in the ablation study to verify the convergence. When all the models, $\phi_1, ..., \phi_n, \phi_b$ are ideally optimized to the same model, they share the same knowledge, thus eliminating information loss in the task-specific model fusion.

\textbf{Meta-training the base model.} In phase 3 of Fig.~\ref{fig:process}, take a small number of samples from all learned tasks to form $\mathbb{M}_n$. $\mathbb{M}_n$ is used for meta-training of $\phi_b^S$ to further optimize the distribution of model parameters. After meta-training, $\phi_b^S$ is the model $\phi_n$ that learned task $n$.

\section{Experiments and Results}
\label{sec:experiment}

We conduct experiments on several common datasets, including MNIST ~\cite{zenke2017continual}, CIFAR100~\cite{krizhevsky2009learning}, CUB200~\cite{wah2011caltech} and ImageNet100 which is a subset of ISLVRC 2012~\cite{russakovsky2015imagenet}. We also perform ablation study to analyze different components of our approach.

\subsection{Datasets}

\textbf{MNIST.} MNIST contains 60k images of handwritten numbers in the training set and 10k samples in the test set. All the images are 28$\times$28 pixels. In our experiment, MNIST is divided into 5 tasks with 2 classes per task. 

\textbf{CIFAR100.} CIFAR100 consists of 60k pictures of 32$\times$32 color images from 100 classes. Each class has 500 training and 100 testing samples. 100 classes are split into 10 tasks with 10 classes in each task.

\textbf{CUB200.} CUB200 contains 200 classes of birds with 11,788 images in total. The training set and the test set consist of 5994 and 5794 images, respectively. The 200 bird classes are split into 6 tasks in our experiment.

\textbf{ImageNet100.} ImageNet100, as a subset of ILSVRC2012, contains 100 classes and 130 thousand samples of 224$\times$224 color images. Each class has about 1,300 training and 50 test samples. We split ImageNet100 into 10 tasks.

\subsection{Implementation Details}

\textbf{Network architecture.} For MNIST, a two-layer MLP is selected as the model. For CIFAR100 and CUB200, the network is ($\mathit{ResNet-18(1/3)}$) which is a reduced version of ResNet-18. For ImageNet100, the original ResNet-18 is used in the experiment. All the architectures used are added the spatial attention mechanism after the first layer.

\textbf{Training details.} For MNIST, each incremental training has 20 epochs. The initial learning rate is set to 0.1 and reduced to 1/2 of the previous learning rate after 5, 10, and 15 epochs. The weight decay is set to 0, the batch size is 256, and $\gamma = 0.1$. The optimizer is SGD. 

For CIFAR100, each incremental training has 70 epochs. The initial learning rate starts from 0.01 and is reduced to 1/5 of the previous learning rate after 30 and 60 epochs. The weight decay is set to 0, the batch size is 512, and $\gamma = 0.1$. The optimizer is set to RAdam\cite{rajasegaran2020itaml}. 

For CUB200 and ImageNet100, each incremental task is trained for 100 epochs. The learning rate starts from 0.1 initially and is reduced to 1/10 of the previous learning rate after 40, 70, and 90 epochs. The weight decay is set to 0, the batch size is 512, and $\gamma = 0.1$. The optimizer is RAdam\cite{rajasegaran2020itaml}. 

For a fair comparison, we set the rehearsal memory size as 2,000 for MNIST and CIFAR100. For CUB200 and Imagenet100, the memory size is set as 3000. The perturbation bound $\epsilon=\frac{8}{255}$ and step size of $\frac{2}{255}$ is set for all the benchmarks.

\subsection{Results and Comparison}

In this section, we report the results on MNIST, CIFAR100, CUB200 and ImageNet100, and compare our ARI method with the state-of-the-art methods.

\begin{figure}[htbp]
\begin{minipage}[]{0.45\linewidth}
\centering
\includegraphics[width=1\linewidth]{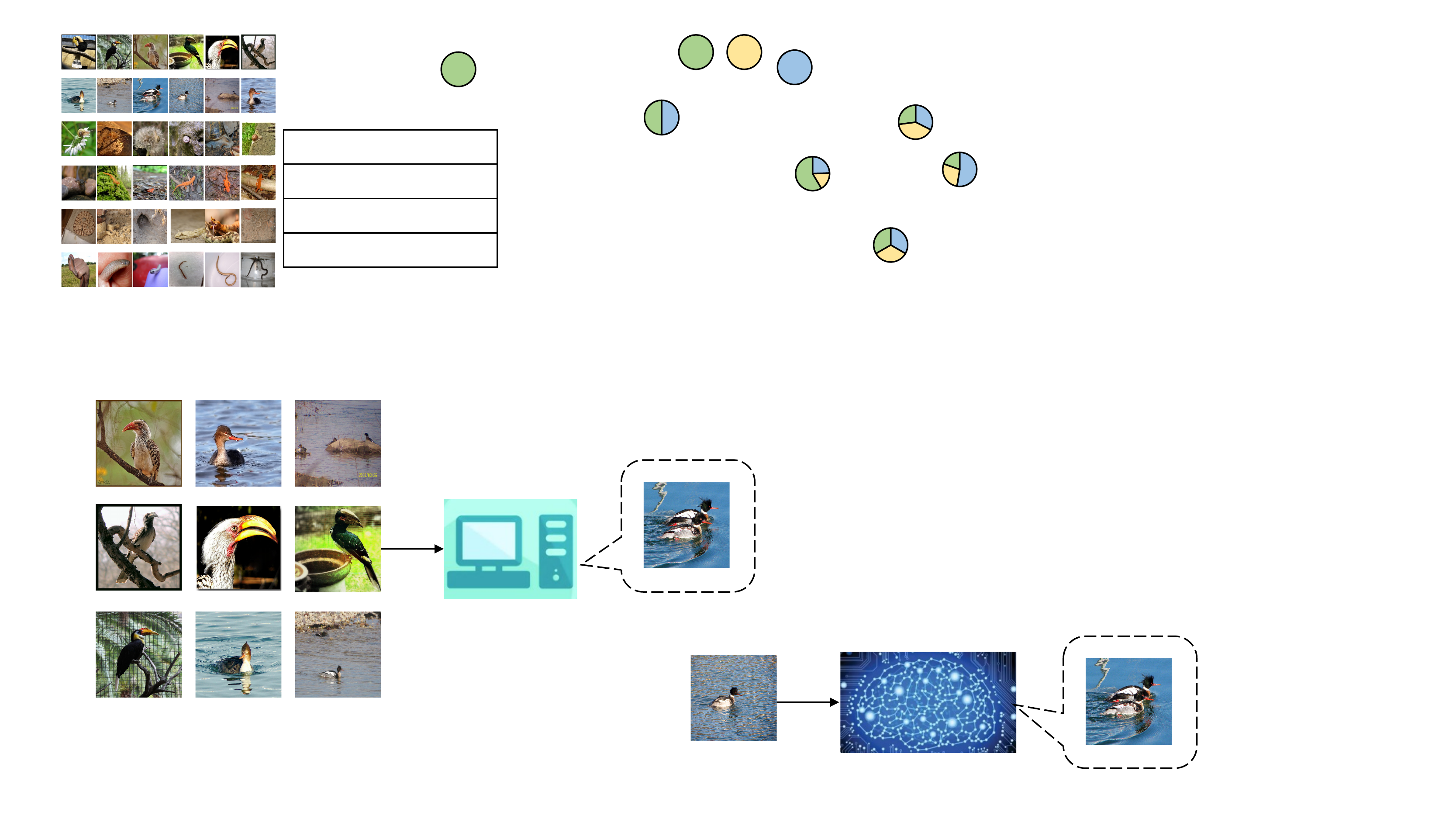}
\caption{Comparison results on the MNIST dataset. ``+" indicates that the method is memory-based.}
\label{fig:MNIST}
\end{minipage}%
\hfill
\begin{minipage}[]{0.45\linewidth}
\centering
\includegraphics[width=1\linewidth]{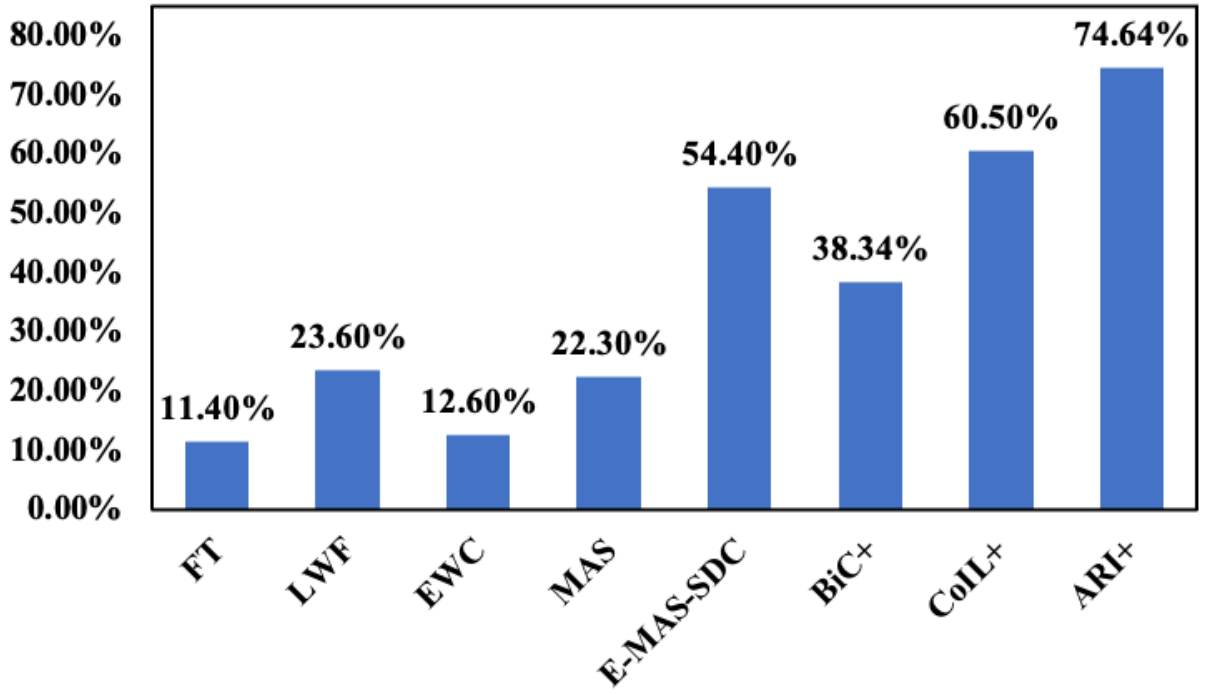}
\caption{The average classiﬁcation accuracy on CUB200, with 6 tasks learned incrementally. ``+" indicates that the method is memory-based.}
\label{fig:result}
\end{minipage}
\end{figure}

\textbf{Small Scale.} The compared typical lifelong learning approaches include Memory Aware Synapses (MAS)~\cite{aljundi2018memory}, LwF~\cite{li2017learning}, Synaptic Intelligence (SI)~\cite{zenke2017continual}, Elastic Weight Consolidation (EWC)~\cite{kirkpatrick2017overcoming},  Gradient Episodic Memory (GEM)~\cite{lopez2017gradient}, Deep Generative  Replay (DGR)~\cite{shin2017continual} and Incremental Task-Agnostic Meta learning (iTAML)~\cite{rajasegaran2020itaml}. As shown in Fig.~\ref{fig:MNIST}, ARI outperforms all the others. Its average classification accuracy of 5 tasks is around 98.91\%.

\textbf{Medium Scale.} ARI attains significant advantages on CIFAR100 compared with other state-of-the-art approaches. For the 10-task lifelong learning, as shown in Table~\ref{tab:CIFAR100} ARI achieves the classification accuracy of 80.88\% which surpasses all the previous methods.

\begin{table*}[ht]
\centering
\caption{Comparison among different lifelong learning methods on CIFAR100. The accuracy of task $t$ is the average accuracy of all $1,2,..,t$ tasks.}
\resizebox{1\columnwidth}{!}{
\begin{tabular}{cccccccccccc}
\toprule
Dataset   & {Methods} & {Task 1}       & {Task 2}       & {Task 3}       & {Task 4}       & {Task 5}       & {Task 6}       & {Task 7}       & {Task 8}       & {Task 9}       & {Task 10}                            \\ \toprule
\multirow{9}{*}{CIFAR100} & {DMC\cite{zhang2020class}}     & {88.11\%} & {76.30\%} & {67.53\%} & {62.19\%} & {57.85\%} & {52.87\%} & {48.59\%} & {43.88\%} & {40.32\%} & {36.28\%}  \\  & {LwF~\cite{li2017learning}}     & {89.30\%} & {70.13\%} & {54.25\%} & {45.78\%} & {39.83\%} & {36.08\%} & {31.67\%} & {28.86\%} & {24.37\%} & {23.86\%}                      \\
                           & {SI~\cite{zenke2017continual}}      & {88.85\%} & {51.76\%} & {40.35\%} & {33.66\%} & {32.01\%} & {29.87\%} & {27.71\%} & {25.97\%} & {24.31}   & {23.51\%}                    \\
                           & {EWC~\cite{kirkpatrick2017overcoming}}     & {88.98\%} & {52.37\%} & {48.37\%} & {38.26\%} & {31.64\%} & {26.14\%} & {21.88\%} & {19.94\%} & {18.76\%} & {16.03\%}                   \\
                           & {MAS~\cite{aljundi2018memory}}     & {88.16\%} & {42.31\%} & {36.16\%} & {35.89\%} & {33.29\%} & {25.97\%} & {21.77\%} & {18.84\%} & {18.11\%} & {15.86\%}                     \\
                           & {RWalk~\cite{chaudhry2018riemannian}}   & {89.57\%} & {55.12\%} & {40.19\%} & {32.54\%} & {29.13\%} & {25.89\%} & {23.61\%} & {21.84\%} & {19.32\%} & {17.91\%}                          \\
                           & {iCARL~\cite{rebuffi2017icarl}}   & {88.74\%} & {78.13\%} & {72.39\%} & {67.23\%} & {63.69\%} & {60.18\%} & {56.35\%} & {54.38\%} & {51.87\%} & {49.46\%} \\
                                                      & {Bic~\cite{wu2019large}}   & {-} & {84.70\%} & {-} & {71.60\%} & {-} & {63.68\%} & {-} & {58.12\%} & {-} & {53.74\%} \\
                           
                           & {iTAML~\cite{rajasegaran2020itaml}}   & \textbf{89.15\%} & \textbf{89.03}\% & \textbf{87.32}\% & \textbf{86.18\%} & \textbf{84.31\%} & \textbf{82.12\%} & {80.65\%} & {79.06\%} & {78.42\%} & {77.79\%} 
                           \\\hline
                        
                           & {ARI}     & {88.60\%} & {86.90\%} & {85.77\%} & {84.55\%} & {83.10\%} & {81.75\%} & \textbf{81.57\%} & \textbf{80.98\%} & \textbf{80.20\%} & \textbf{80.88\%}                           \\ \bottomrule
\end{tabular}}
\label{tab:CIFAR100}
\end{table*}

We calculate the metrics BWT~\cite{lopez2017gradient} and FWT~\cite{lopez2017gradient} to measure forgetting and learning. As shown in Table~\ref{tab:bwt}, although the BWT value of GEM is the highest, its accuracy (65.4\%) is much lower than ours (80.88\%). As mentioned in \cite{lopez2017gradient}, the BWT and FWT of two methods can indicate their performances only when they have similar accuracies.

\begin{table}[h]
    \centering
    \caption{Comparison of forgetting metrics on CIFAR100.}
    \resizebox{0.8\columnwidth}{!}{
    \begin{tabular}{ccccccc}
    \toprule
    \textbf{}    & UCIR   & GEM   & PODNet\cite{douillard2020podnet}     & iTAML   & iTAML+RRR & ARI \\ \toprule
    BWT & -8.5\% & \textbf{1.2}\% & -16.3\% & -11.5\% & -8.5\%    & -7.5\%  \\
    FWT & -5.56\% & 0.47\% & -5.58\% & 0.14\% & 0.77\%    & \textbf{1.18\%}  \\ \toprule
    \end{tabular}}
    \label{tab:bwt}
\end{table}

We also evaluate ARI's efficiency on CIFAR100. The memory complexity is similar to other memory-based methods. Its extra memory addition is the dictionary to hold the parameters of the specific models. This extra memory is only about 100MB, which is negligible compared with the memory requirement during training (7200MB). Its time complexity increases by 20\% due to the background adversary. The CIFAR100 experiment takes 3.3 hours by one TITAN XP when total epochs=70 and batch size=512.

In Fig.~\ref{fig:result}, We compare different methods on CUB200 with 6 incremental tasks where BiC~\cite{wu2019large} and CoIL~\cite{zhou2021co} are memory-based methods.  ARI surpasses CoIL by 14.14\%, which illustrates that ARI is less prone to catastrophic forgetting.

\textbf{Large Scale.} We compare ARI with the state-of-the-art algorithms on the large scale dataset ImageNet100. The comparison results are listed in Table~\ref{tab:imagenet}, where Mem\% denotes the proportion of the memory size $\mathbb{M}$ in the Imagenet100 training set. ARI outperforms Fixed representations (FixedRep) and other methods of lifelong learning. ARI increases the accuracy on ImageNet100 by $5.22\%$ (from $74.10\%$ to $79.32\%$).

\begin{table}[]
\centering
\caption{Comparison of different approaches on ImageNet100.}
\resizebox{0.55\columnwidth}{!}{\begin{tabular}{cccc}
\toprule
\multicolumn{1}{l}{Datasets}  & Methods & Accuracy &Mem\%\\ \toprule
\multirow{5}{*}{ImageNet100} 

    & iCaRL~\cite{rebuffi2017icarl}   & 63.50\% &2\% \\
    & UCIR~\cite{hou2019learning}    & 69.09\%  &2\%\\
    & MARK~\cite{hurtado2021optimizing}    & 69.43\%  &10\%\\
    &DER~\cite{yan2021dynamically}    & 66.70\% &2\% \\
    &RPSnet~\cite{rajasegaran2019random}    & 74.10\% &2\% \\
    \hline
                             
                & ARI    & \textbf{79.32\%} &3\%   \\ 
                         \bottomrule
\end{tabular}}
\label{tab:imagenet}
\end{table}
The above results illustrate ARI's consistent effectiveness and superiority on small, medium, and large scale datasets over other methods.

\subsection{Ablation Study}
\label{sec:4.4}

In this section, we conduct extensive experiments to verify the effects of the proposed background attack and task-specific model fusion.

\begin{figure}[h]
    \centering
    \includegraphics[width=0.95\linewidth]{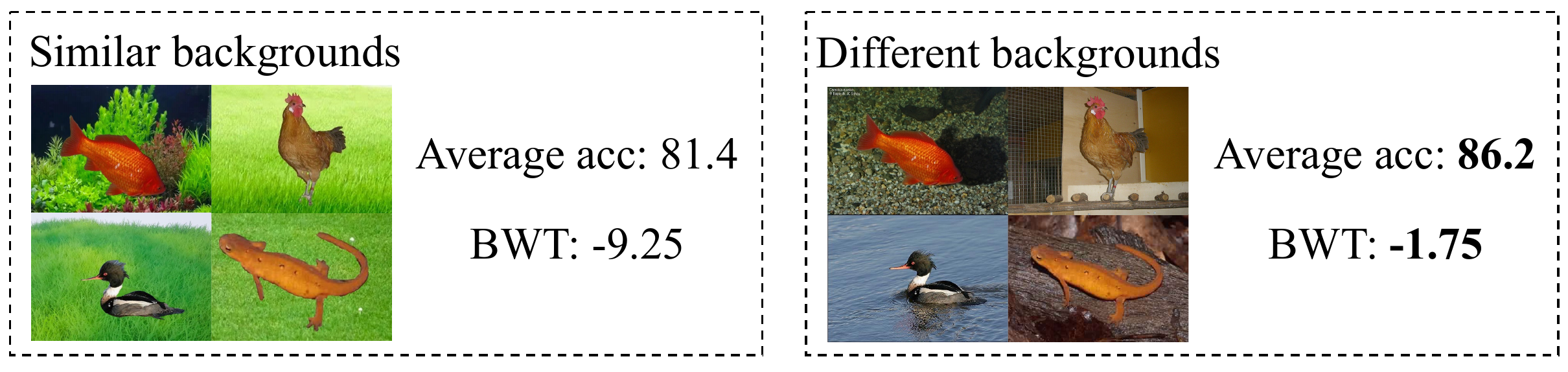}
    \caption{Similar characteristics lead to the forgetting of lifelong learning.}
    \label{fig:similar_back}
\end{figure}

\textbf{Similar characteristics lead to forgetting.} First we provide a toy experiment to demonstrate that retroactive interference leads to forgetting, as shown in Fig.~\ref{fig:similar_back}. We construct a dataset with 10 categories, each containing 100 training and 50 test images. We replace the training image
backgrounds with similar backgrounds but do not change the test images. We form 5 tasks, each with 2 categories. Average accuracy and BWT are evaluated after all tasks are trained, and the result is compared with its counterpart from the original training images with different backgrounds. The results show that similar characteristics cause forgetting.

\begin{figure}[htbp]
\begin{minipage}[]{0.47\linewidth}
\centering
\includegraphics[width=1\linewidth]{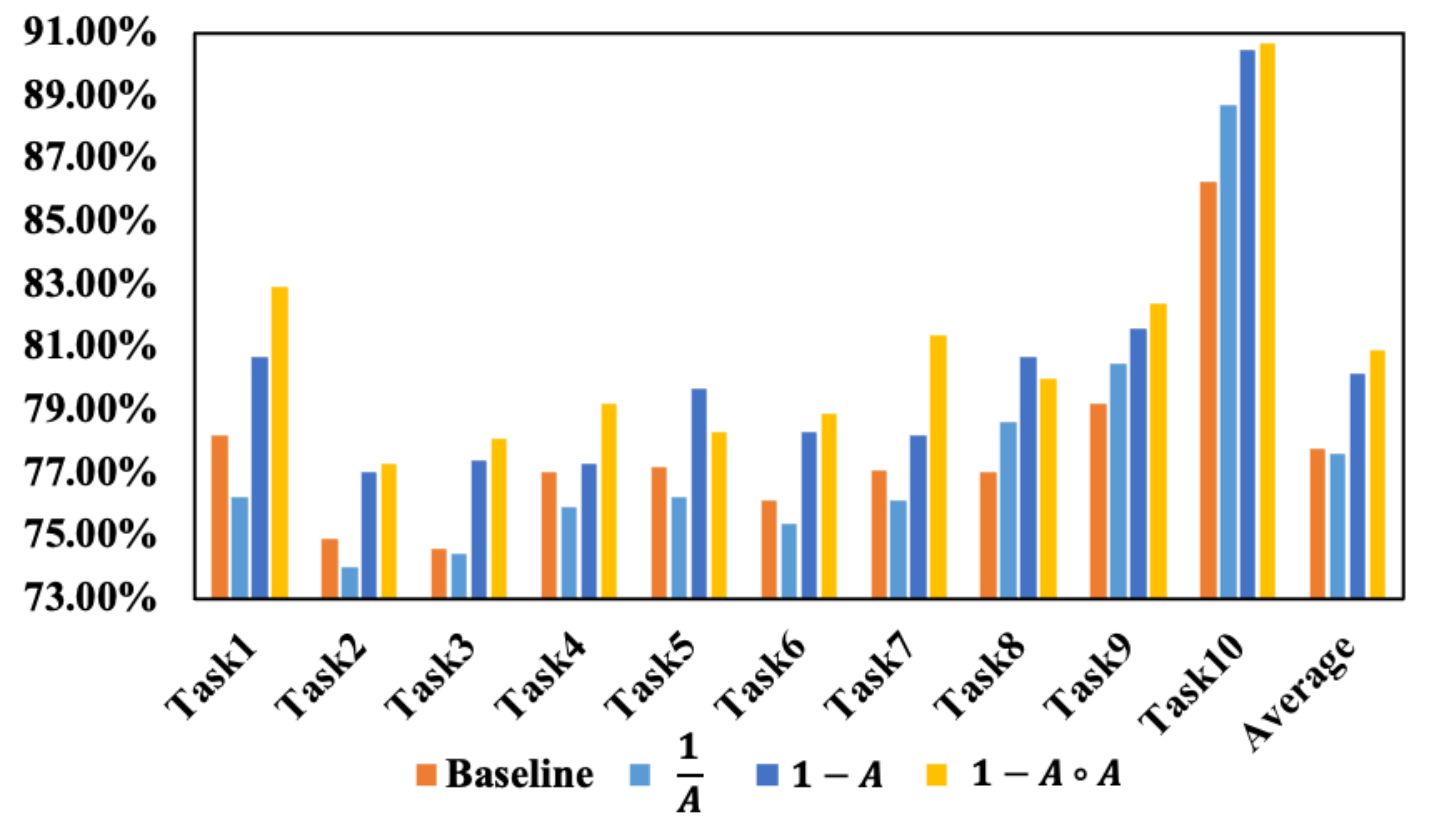}
\caption{The impacts of various types of background mask on the lifelong learning process on CIFAR100. The baseline is the model without the attack.} 
\label{fig:a} 
\end{minipage}%
\hfill
\begin{minipage}[]{0.5\linewidth}
\centering
\includegraphics[width=1\linewidth]{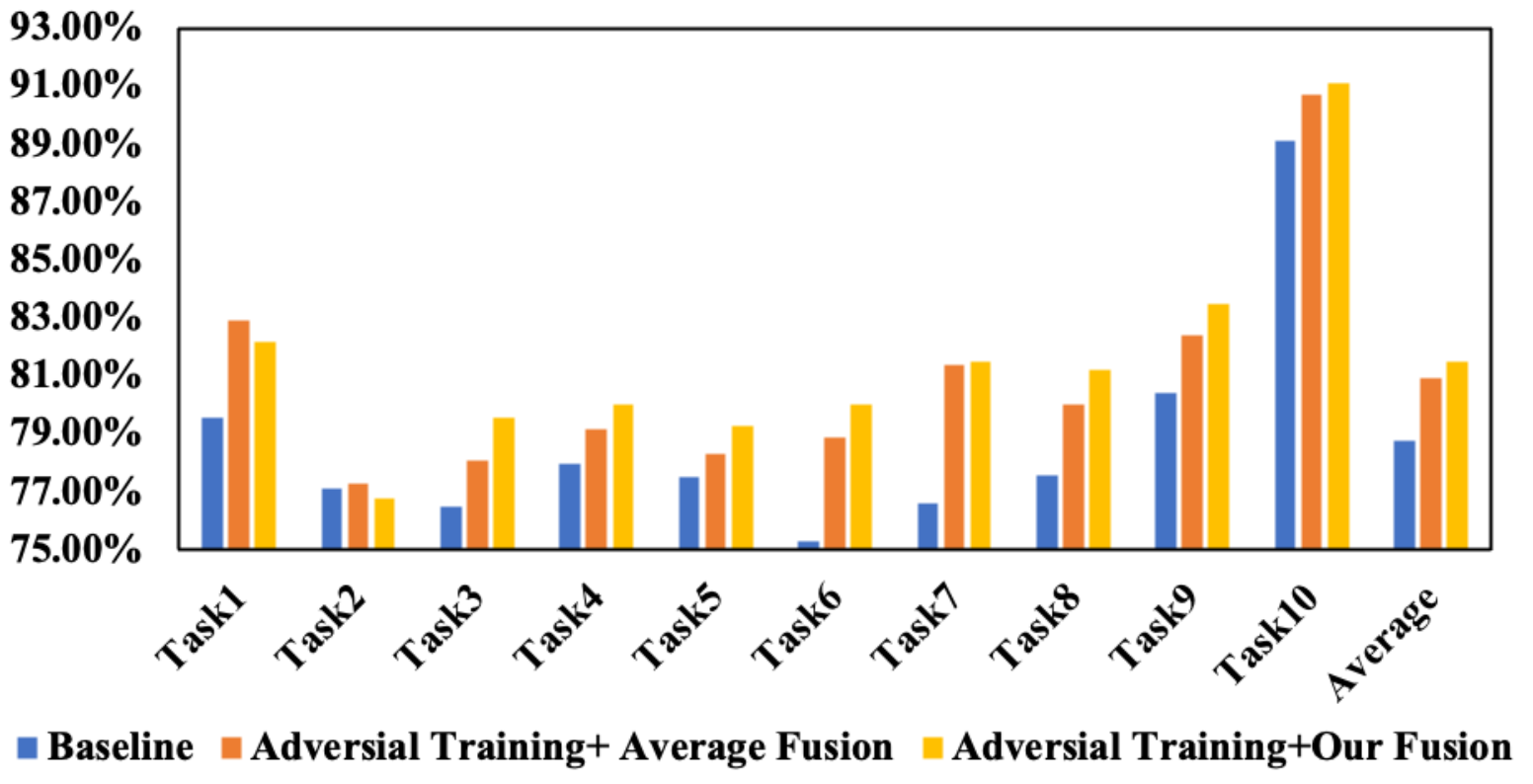}
\caption{The classification accuracy in lifelong learning process on CIFAR100. The baseline is the same model but without the attack and with average model fusion.} 
\label{fig:b}
\end{minipage}
\end{figure}

\begin{figure}[htbp]
    \centering
	\begin{minipage}{0.5\linewidth}
		\centering
        \captionof{table}{We conduct 5 tests and analyze the data distribution of $\mathbf{A}$. \textit{std} denotes the standard deviation. The majority of the values are closer to $1$ than to $0$.}
        \resizebox{1\columnwidth}{!}{
        \begin{tabular}{cccccc}
        \toprule
                  & test1 & test2 & test3 & test4 & test5 \\ \hline
        $\left[0,0.3\right)$   & 2.03\%  & 1.19\%  & 4.22\%  & 2.62\%  & 1.72\%  \\
        $\left[0.3,0.5\right)$  & 7.62\%  & 5.86\%  & 17.43\% & 12.62\% & 4.65\%  \\
        $\left[0.5,0.7\right)$  & 75.03\% & 29.42\% & 59.59\% & 55.43\% & 81.93\% \\
        $\left[0.7,1\right]$    & 15.32\% & 63.54\% & 18.76\% & 29.33\% & 11.70\% \\
        std    & 14.35\% & 13.72\% & 17.48\% & 15.14\% & 14.52\% \\ \bottomrule
        \end{tabular}}
        \label{tab:kernel}
	\end{minipage}
	\hfill	
	\begin{minipage}{0.45\linewidth}
		\centering
        \includegraphics[width=1\linewidth]{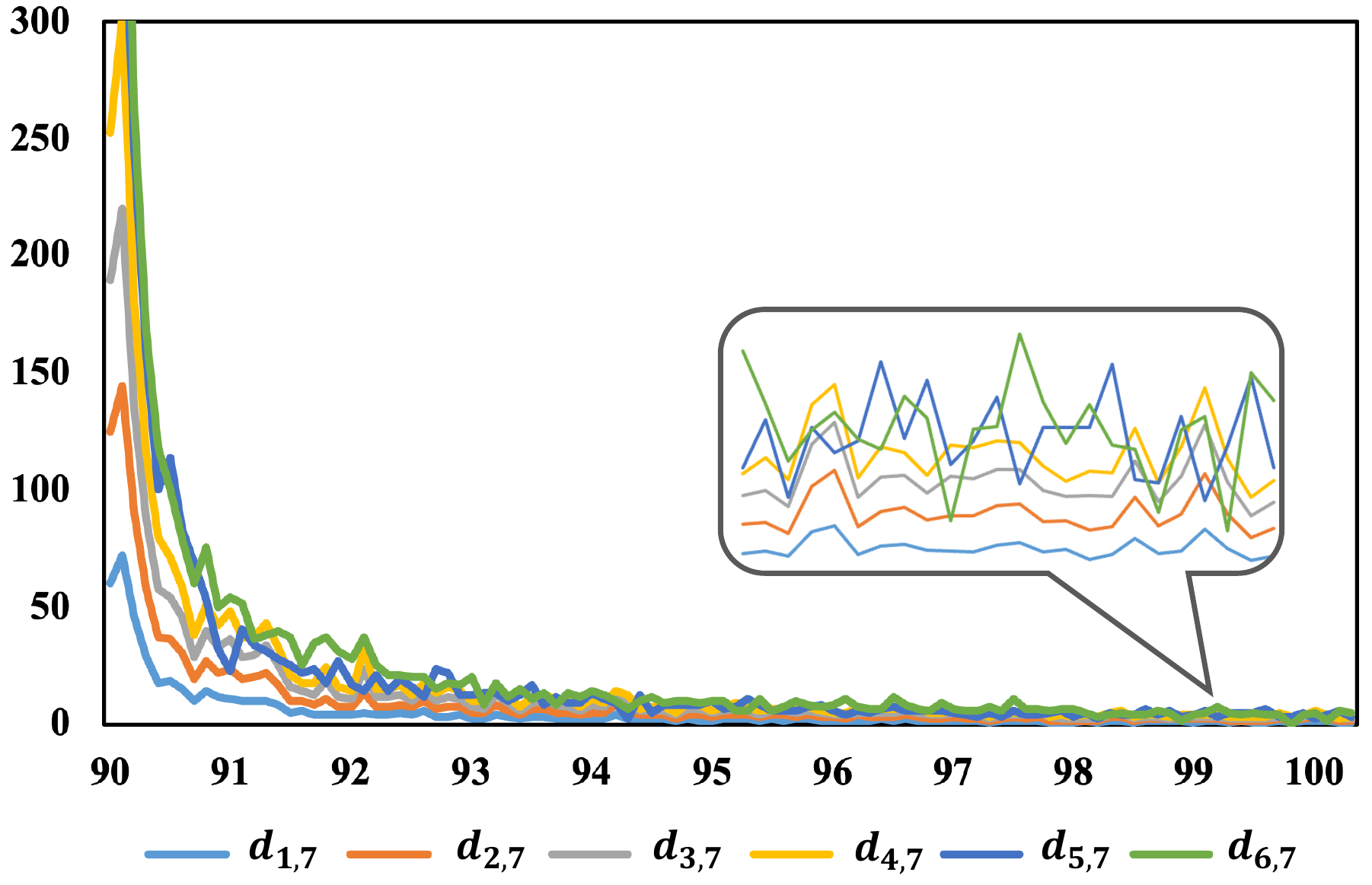}
    	\caption{The distance between task-specific models and base model on CUB200.  } 
    	\label{fig:dif_cub}
	\end{minipage}
\end{figure}

\textbf{The effects of different $\mathbf{B}$.} During the adversarial training process, we apply different ways to make background attacks. We set the mask $\mathbf{B}$ as $(\frac{1}{\mathbf{A}})$, $(1-\mathbf{A})$ and $(1-\mathbf{A}\circ\mathbf{A})$, and perform experiments respectively. As shown in Fig.~\ref{fig:a}, the performance varies according to the background mask $\mathbf{B}$. The results performed by $(1-\mathbf{A}\circ\mathbf{A})$ are better than other methods. In order to analyze the cause of the various impact, we randomly sample 5 attention masks and list their distributions in Table~\ref{tab:kernel}. Since the masks have values close to $0$, the attack with $\mathbf{B}=(\frac{1}{\mathbf{A}})$ would be so huge that it decreases the robustness of the model. Moreover, because the values are closer to $1$ than to $0$, $\mathbf{B}=(1-\mathbf{A}\circ\mathbf{A})$ can widen the distance between foreground and background more effectively than $\mathbf{B}=1-\mathbf{A}$. In our experiments, the attack using $\mathbf{B}=(1-\mathbf{A}\circ\mathbf{A})$ performs the best, meaning that it can guide background attack more effectively. The visualization results of $\mathbf{A}$ and $\mathbf{B}$ are presented in the supplementary material.

\textbf{The effects of model fusion.} To verify the effect of our proposed model fusion method, we compare the lifelong learning results with and without our model fusion on CIFAR100 while keeping the other settings unchanged. The results are shown in Fig.~\ref{fig:b}. It could be observed that our fusion operation makes the learning better incrementally. To verify whether the task-specific models tend to be similar, we test the values of $\mathbf{dif}$. In Fig.~\ref{fig:dif_cub}, we intercept the 90 -- 100 epochs on the CUB200 benchmark. The task number $n$ equals $6$ as shown in Eq.~\ref{eq:cor}. The vertical axis represents the distances between task-specific models and the base model. As the training progresses, the distance gradually converges to 0. Through our model fusion method, different task-specific models can converge to the optimal one, thus eliminating information loss and retroactive interference in the task-specific model fusion, which illustrates the effectiveness of our method.

\section{Conclusion}

Lifelong learning aims to learn a single model that can continuously adapt to the new knowledge without overriding existing knowledge. 
We develop a meta-learning approach to train a base model which can be efficiently optimized for lifelong learning. First, a background attack method is introduced to extract critical features and avoid retroactive interference. Then, an adaptive weight fusion mechanism is presented according to the distances between the base and the task-specific models. Our experiments demonstrate consistent improvements across a range of classiﬁcation datasets, including ImageNet100, CUB200, CIFAR100, and MNIST.

\section*{Acknowledgements}
This work was supported in part by the National Natural Science Foundation of China under Grant 62076016. We gratefully acknowledge the support of MindSpore, CANN (Compute Architecture for Neural Networks) and Ascend AI processor used for this research \footnote{MindSpore. https://www.mindspore.cn/}.

\bibliographystyle{splncs04}
\bibliography{egbib}

\begin{thebibliography}{10}
\providecommand{\url}[1]{\texttt{#1}}
\providecommand{\urlprefix}{URL }
\providecommand{\doi}[1]{https://doi.org/#1}

\bibitem{abati2020conditional}
Abati, D., Tomczak, J., Blankevoort, T., Calderara, S., Cucchiara, R.,
  Bejnordi, B.E.: Conditional channel gated networks for task-aware continual
  learning. In: CVPR (2020)

\bibitem{aljundi2018memory}
Aljundi, R., Babiloni, F., Elhoseiny, M., Rohrbach, M., Tuytelaars, T.: Memory
  aware synapses: Learning what (not) to forget. In: ECCV (2018)

\bibitem{ausubel1961role}
Ausubel, D.P., Fitzgerald, D.: The role of discriminability in meaningful
  learning and retention. Journal of Educational Psychology  \textbf{52}(5),
  ~266 (1961)

\bibitem{castro2018end}
Castro, F.M., Mar{\'\i}n-Jim{\'e}nez, M.J., Guil, N., Schmid, C., Alahari, K.:
  End-to-end incremental learning. In: ECCV (2018)

\bibitem{chaudhry2018riemannian}
Chaudhry, A., Dokania, P.K., Ajanthan, T., Torr, P.H.: Riemannian walk for
  incremental learning: Understanding forgetting and intransigence. In: ECCV
  (2018)

\bibitem{douillard2020podnet}
Douillard, A., Cord, M., Ollion, C., Robert, T., Valle, E.: Podnet: Pooled
  outputs distillation for small-tasks incremental learning. In: ECCV (2020)

\bibitem{Ebrahimi2021RememberingFT}
Ebrahimi, S., Petryk, S., Gokul, A., Gan, W., Gonzalez, J., Rohrbach, M.,
  Darrell, T.: Remembering for the right reasons: Explanations reduce
  catastrophic forgetting. In: ICLR (2021)

\bibitem{goodfellow2014explaining}
Goodfellow, I.J., Shlens, J., Szegedy, C.: Explaining and harnessing
  adversarial examples. In: ICLR (2015)

\bibitem{goodfellow2016deep}
Goodfellow, I., Bengio, Y., Courville, A.: Deep Learning. MIT press (2016)

\bibitem{hadsell2020embracing}
Hadsell, R., Rao, D., Rusu, A.A., Pascanu, R.: Embracing change: Continual
  learning in deep neural networks. Trends in Cognitive Sciences  (2020)

\bibitem{hou2019learning}
Hou, S., Pan, X., Loy, C.C., Wang, Z., Lin, D.: Learning a unified classifier
  incrementally via rebalancing. In: CVPR (2019)

\bibitem{hurtado2021optimizing}
Hurtado, J., Raymond-Saez, A., Soto, A.: Optimizing reusable knowledge for
  continual learning via metalearning. In: NeurIPS (2021)

\bibitem{kirkpatrick2017overcoming}
Kirkpatrick, J., Pascanu, R., Rabinowitz, N., Veness, J., Desjardins, G., Rusu,
  A.A., Milan, K., Quan, J., Ramalho, T., Grabska-Barwinska, A., et~al.:
  Overcoming catastrophic forgetting in neural networks. Proceedings of the
  National Academy of Sciences  \textbf{114}(13),  3521--3526 (2017)

\bibitem{krizhevsky2009learning}
Krizhevsky, A., Hinton, G., et~al.: Learning multiple layers of features from
  tiny images. Tech. rep. (2009)

\bibitem{kurakin2016adversarial}
Kurakin, A., Goodfellow, I., Bengio, S.: Adversarial examples in the physical
  world. In: ICLR (2016)

\bibitem{li2017learning}
Li, Z., Hoiem, D.: Learning without forgetting. IEEE Transactions on Pattern
  Analysis and Machine Intelligence  \textbf{40}(12),  2935--2947 (2017)

\bibitem{lopez2017gradient}
Lopez-Paz, D., Ranzato, M.: Gradient episodic memory for continual learning.
  In: NeurIPS (2017)

\bibitem{na2017cascade}
Na, T., Ko, J.H., Mukhopadhyay, S.: Cascade adversarial machine learning
  regularized with a unified embedding. In: ICLR (2017)

\bibitem{mrmr}
Peng~H, Long~F, D.C.: Feature selection based on mutual information criteria of
  max-dependency, max-relevance, and min-redundancy. IEEE Transactions on
  Pattern Analysis and Machine Intelligence pp. 1226--1238 (2005)

\bibitem{rajasegaran2019random}
Rajasegaran, J., Hayat, M., Khan, S.H., Khan, F.S., Shao, L.: Random path
  selection for continual learning. In: NeurIPS (2019)

\bibitem{rajasegaran2020itaml}
Rajasegaran, J., Khan, S., Hayat, M., Khan, F.S., Shah, M.: itaml: An
  incremental task-agnostic meta-learning approach. In: CVPR (2020)

\bibitem{rebuffi2017icarl}
Rebuffi, S.A., Kolesnikov, A., Sperl, G., Lampert, C.H.: icarl: Incremental
  classifier and representation learning. In: CVPR (2017)

\bibitem{russakovsky2015imagenet}
Russakovsky, O., Deng, J., Su, H., Krause, J., Satheesh, S., Ma, S., Huang, Z.,
  Karpathy, A., Khosla, A., Bernstein, M., et~al.: Imagenet large scale visual
  recognition challenge. International Journal of Computer Vision
  \textbf{115}(3),  211--252 (2015)

\bibitem{shin2017continual}
Shin, H., Lee, J.K., Kim, J., Kim, J.: Continual learning with deep generative
  replay. In: NeurIPS (2017)

\bibitem{sternberg2012cognitive}
Sternberg, R.J., Sternberg, K., Mio, J.: Cognitive psychology. Cengage Learning
  Press (2012)

\bibitem{tramer2018ensemble}
Tram{\`e}r, F., Boneh, D., Kurakin, A., Goodfellow, I., Papernot, N., McDaniel,
  P.: Ensemble adversarial training: Attacks and defenses. In: ICLR (2018)

\bibitem{vogel2005neural}
Vogel, E.K., McCollough, A.W., Machizawa, M.G.: Neural measures reveal
  individual differences in controlling access to working memory. Nature
  \textbf{438}(7067),  500--503 (2005)

\bibitem{wah2011caltech}
Wah, C., Branson, S., Welinder, P., Perona, P., Belongie, S.: The caltech-ucsd
  birds-200. Technical Report CNS-TR-2010-001  (2011)

\bibitem{woo2018cbam}
Woo, S., Park, J., Lee, J.Y., Kweon, I.S.: Cbam: Convolutional block attention
  module. In: ECCV (2018)

\bibitem{wu2019large}
Wu, Y., Chen, Y., Wang, L., Ye, Y., Liu, Z., Guo, Y., Fu, Y.: Large scale
  incremental learning. In: CVPR (2019)

\bibitem{yan2021dynamically}
Yan, S., Xie, J., He, X.: Der: Dynamically expandable representation for class
  incremental learning. In: CVPR (2021)

\bibitem{yu2020semantic}
Yu, L., Twardowski, B., Liu, X., Herranz, L., Wang, K., Cheng, Y., Jui, S.,
  Weijer, J.v.d.: Semantic drift compensation for class-incremental learning.
  In: CVPR (2020)

\bibitem{zenke2017continual}
Zenke, F., Poole, B., Ganguli, S.: Continual learning through synaptic
  intelligence. In: ICML (2017)

\bibitem{zhang2021delving}
Zhang, C., Zhang, M., Zhang, S., Jin, D., Zhou, Q., Cai, Z., Zhao, H., Yi, S.,
  Liu, X., Liu, Z.: Delving deep into the generalization of vision transformers
  under distribution shifts. arXiv  (2021)

\bibitem{zhang2020class}
Zhang, J., Zhang, J., Ghosh, S., Li, D., Tasci, S., Heck, L., Zhang, H., Kuo,
  C.C.J.: Class-incremental learning via deep model consolidation. In: WACV
  (2020)

\bibitem{zhao2020maintaining}
Zhao, B., Xiao, X., Gan, G., Zhang, B., Xia, S.T.: Maintaining discrimination
  and fairness in class incremental learning. In: CVPR (2020)

\bibitem{zhou2021co}
Zhou, D., Ye, H., Zhan, D.: Co-transport for class-incremental learning. In:
  ACM MM (2021)

\end{thebibliography}
\end{document}